\documentclass[letterpaper]{article} % DO NOT CHANGE THIS
\usepackage{aaai24}  % DO NOT CHANGE THIS
\usepackage{times}  % DO NOT CHANGE THIS
\usepackage{helvet}  % DO NOT CHANGE THIS
\usepackage{courier}  % DO NOT CHANGE THIS
\usepackage[hyphens]{url}  % DO NOT CHANGE THIS
\usepackage{graphicx} % DO NOT CHANGE THIS
\usepackage{natbib}  % DO NOT CHANGE THIS AND DO NOT ADD ANY OPTIONS TO IT
\usepackage{caption} % DO NOT CHANGE THIS AND DO NOT ADD ANY OPTIONS TO IT
\usepackage{algorithm}
\usepackage{algorithmic}

\usepackage{amsfonts}

\usepackage{subcaption}

\usepackage{todonotes}
\usepackage{habbas_macros}
\usepackage{ltl}
\usepackage{soul}

\newtheorem{definition}{Definition}
\newcommand{\fullsstit}{\OEsstit{\alpha}{P_{\geq \rho}\varphi}{\pi}}
\usepackage{newfloat}
\usepackage{listings}
\DeclareCaptionStyle{ruled}{labelfont=normalfont,labelsep=colon,strut=off} % DO NOT CHANGE THIS
\floatstyle{ruled}
\newfloat{listing}{tb}{lst}{}
\floatname{listing}{Listing}
\title{Formal Ethical Obligations in Reinforcement Learning Agents: Verification and Policy Updates}
\author{
    Colin Shea-Blymyer\textsuperscript{\rm 1}\footnote{Work done while at Oregon State University}, 
    Houssam Abbas\textsuperscript{\rm 2}
}
\affiliations{
    \textsuperscript{\rm 1}
    Georgetown University, Center for Security and Emerging Technology\\
    colin.sheablymyer@georgetown.edu

    \textsuperscript{\rm 2}
    Oregon State University, School of Electrical Engineering and Computer Science
}

\begin{document}

\maketitle

\begin{abstract}
% \todo[inline]{the term Obligations is not widely understood or clear, so it's hard to understand what you mean by it, yet it's clearly critical to what this paper is about...So unlike usually, delete first sentence and distribute its content in the rest of hte abstract. So start with current 2nd sentence, }
% \st{We develop a deontic logic for reasoning about a reinforcement learning agent's obligations under an optimal policy, and two corresponding algorithms: one for model-checking whether a (large) RL agent has the right strategic obligations, and one for modifying a reference decision policy to make it meet obligations expressed in the logic}.
When designing agents for operation in uncertain environments, designers need tools to automatically reason about what agents ought to do, how that conflicts with what is actually happening, and how a policy might be modified to remove the conflict.
These obligations include ethical and social obligations, permissions and prohibitions, which constrain how the agent achieves its mission and executes its policy.
We propose a new deontic logic, Expected Act Utilitarian deontic logic, for enabling this reasoning at design time: for specifying and verifying the agent's strategic obligations, then modifying its policy from a reference policy to meet those obligations.
Unlike approaches that work at the reward level, working at the logical level increases the transparency of the trade-offs.
We introduce two algorithms: one for model-checking whether an RL agent has the right strategic obligations, and one for modifying a reference decision policy to make it meet obligations expressed in our logic.
We illustrate our algorithms on DAC-MDPs which accurately abstract neural decision policies, and on toy gridworld environments.
\end{abstract}

\section{Introduction: Strategic Obligations in the Face of Uncertainty}
% Reinforcement Learning has been widely adopted to train and deploy autonomous agents within complex and often unpredicatable environments. [cit here]. While much effort has been put into increasing the performance of these agents in the explicit reward function, a critical dimension of how these agents align implicit expectations (such as alignment with societal, ethical and normativ expectations) remains under-explored. Given the increasing ubiquity of RL agents in real-world scenarios, from finance to healthcare to transportation, it is essential to establish mechanisms that ensure their actions do not merely optimize for a primary objective but also adhere to a broader spectrum of implicit expectations, including ethical norms and societal standards. For brevity, we refer to these ethical and social standards as obligations.

In the rapidly evolving domain of reinforcement learning (RL), agents are trained to autonomously perform tasks within complex and often unpredictable environments. 
While significant advancements have been made in improving the performance and adaptability of these agents, a crucial dimension remains under-explored: how these agents align with societal and ethical expectations.
Given the increasing ubiquity of RL agents in real-world scenarios — from finance \cite{hambly2023recent} to health care \cite{yu2021reinforcement} and transportation \cite{sallab2017deep} — it is essential to establish mechanisms that ensure their actions do not merely optimize for the mission's objective but also adhere to a broader spectrum of ethical norms and societal standards. For brevity, we refer to these ethical and social standards as \textit{obligations}.
Without such normative constraints, the agent's behavior is likely to be badly surprising, and ultimately unsafe when we think of the reactions of human agents in the environment.

\paragraph{Running example.} 

\begin{figure}[t]
    \centering
    \includegraphics[width=\linewidth]{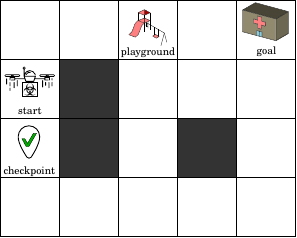}
    \caption{\yhl{The windy-drone MDP. Darkened cells are inaccessible states. The goal state is an absorbing state.} An agent in \yhl{this MDP} has 4 actions available to it in any state: up, down, left, and right. A chosen action has a 70\% chance of success, and on a failure the agent ``slips'' in one of the unchosen directions. An action result that would move the agent into a wall, or other inaccessible state, leaves the agent in the state it acted from.}
    \label{fig:windydrone}
\end{figure}

Figure \ref{fig:windydrone} shows a problem where a drone must carry biohazardous material across a city to a hospital. The RL problem is to maximize the drone's expected utility, and the rewards are assigned such that the hospital rewards the drone with 10 points, but every other space gives the drone a penalty of -1. The penalty is designed to reflect the operating cost of the drone, and the opportunity cost of not being at its goal. The reward reflects the utility of the drone arriving and remaining at the hospital. This reward encodes the drone's \textit{mission}. Solving this RL problem gives the policy that leads to the shortest route to the hospital.

However, we also want the drone to avoid a children's playground, where accidental contamination is especially problematic. This indicates a moral conflict between delivering the material to hospital patients quickly, and avoiding contamination to third parties. Instead of tweaking the reward or designing a secondary reward, we assign the drone an explicit, legible, logical obligation to avoid the playground with a high probability. The requirement is probabilistic to account for the uncertainty in the environment, which could make a non-probabilistic obligation unachievable in every case.
This explicit obligation, and the algorithms we develop to handle it, avoid the pitfalls of reward manipulation, especially as the number of moral dilemmas increases, and the reward balancing potentially becomes more and more arbitrary.
% ~$\square$

For this task we need ways to \textit{formalize} the right obligations, and tools to guarantee that the agent has adopted a policy that maximizes its utility only subject to meeting these obligations.
Traditional specification languages, like Linear Temporal Logic (LTL), are inadequate at drawing a distinction between what is the case (how the agent behaves) and what should be the case (how it \textit{should} behave to meet its obligations). 
Instead, a \textit{deontic logic} is needed for this distinction~\cite{McNamaraChapter}.
We adopt the logic of Expected Act Utilitarianism (EAU), first introduced in \cite{shea2022eau}.
EAU allows specification and automatic reasoning about the obligations of utility maximizing systems, including the models that underlie reinforcement learning. But it has the shortcoming of grounding obligations in the agent's optimal action at a given moment, without regard for future actions, which conflicts with the way a policy is computed as the maximizer of \textit{long-term} utility.
To remedy this, we introduce a \textit{strategic} modality to EAU which grounds obligations in the agent's entire policy, not just instantaneous action.

Designers also need algorithms that can both verify if a given agent's policy meets the specified obligations, and guide modifications to the agent's policy if it falls short.
This paper introduces two algorithms for these challenges: the first efficiently model-checks an RL agent's policy against the strategic obligations formalized in EAU, and the second employs policy gradient ascent to refine the agent's policy until it aligns with the given obligations.

This paper's contributions are:
\begin{itemize}
    \item An extension of EAU to strategic obligations, which formalize obligations that must be met by the optimal (utility-maximizing) policy over an infinite horizon, not just the current time step (Section~\ref{sec:mc}).
    \item An algorithm for model-checking whether a Markov Decision Process (MDP), which models action under uncertainty, has given strategic obligations formalized in (the extended)  EAU. The model-checker can handle MDPs with tens of thousands of states (Section~\ref{sec:mc}).
    \item An algorithm for modifying a utility-maximizing policy to also satisfy a deontic obligation (Section~\ref{sec:policy update}).
    \item An extension of the above policy search algorithm for the case where rewards are not known \textit{a priori} (Section~\ref{sec:experiments}).
    \item Experimental evidence for the effectiveness of the model checking and policy search algorithms (Section~\ref{sec:experiments}).
\end{itemize}

% \begin{quote}\begin{scriptsize}\begin{verbatim}
% \documentclass[letterpaper]{article}
% \usepackage[submission]{aaai24}
% \end{verbatim}\end{scriptsize}\end{quote}

%=============================================
\section{Related Work}
\label{sec:related}
There are several works that propose approaches for developing ethical RL agents.
In \cite{Abel2016ReinforcementLA}, the authors argue that RL provides a sufficient foundation to ground ethical decision making, and \cite{gerdes2015implementable} explores methods for implementing ethics into RL systems.
Works such as \cite{Noothigattu2019TeachingAA} and \cite{Wu2017ALE} propose the use of inverse reinforcement learning to teach norms to agents.
However \cite{Arnold2017ValueAO} argues for a hybrid approach, and \cite{Bringsjord2006TowardAG} offers a purely logical approach.
In this paper, we aim to maintain the precision and formality of a logical approach while extending it to be compatible with common RL techniques.

% \subsection{Policy Gradient Methods}
% \label{sec:pgm}
Our algorithm to modify an agent's policy resembles work in policy gradient methods \cite{sutton2000comparing}.
Policy gradient methods are robust and versatile mechanisms within the realm of on-policy reinforcement learning and have been used in safe reinforcement learning \cite{gu2022review}. These techniques are designed to optimize cumulative reward by directly manipulating policy functions, affording ease of implementation and compatibility with function approximations. The foundational work of REINFORCE \cite{Williams1992SimpleSG} marked an early instance where policy gradients were computed using Monte Carlo returns. 
However, our approach relies entirely on the dynamics of the system, and is concerned both with cumulative reward \textit{and} conformance to an obligation.
% \todo[inline]{start by saying why you're talking about this. E.g. 'We use policy gradient methods to do X'. Also this reads like a related works section, there is nothing technical in it. Do you need to describe something more closely?}

% \tdhl{Connection to CMDPs, ``verifiably safe off-model RL'', etc.}
\yhl{
This reliance on system dynamics makes our problem similar to the problem addressed in \cite{wolf2012robustcontrol}.
There, the authors produce an MDP that encodes  logical constraints (in a non-deontic logic) and can be solved with dynamic programming.
However, their solution maximizes the probability of satisfying the logical constraint while ours seeks to maximize expected utility subject to satisfying a logical constraint.
}

\yhl{
Our goal of maximizing expected utility subject to the satisfaction of an obligation is analogous to the constrained MDP (CMDP) problem \cite{altman2021constrained}.
The CMDP problem is to find the policy that maximizes expected utility subject to a constraint on expected cost (or secondary reward).
% Our second experiment in Section \ref{sec:illustrative} reflects this problem formulation, but
We seek to constrain the MDP solution with a non-discounted probability of reaching a state instead of with a discounted cost.
}
The non-discounted nature of the probability of reaching a state makes our problem distinct from the CMDP problem.
Further, the model checkers we employ do not consider discounted rewards.
Thus, our problem is not solved by solutions to the CMDP problem or by solutions to a classical model checking problem.
% \todo[inline]{more detail on the difference between our problem and the CMDP problem. Maybe also mention that discounted rewards aren't handled by, e.g. STORM - or put that in the approaches section\dots}

%=============================================
\section{Technical Preliminaries}
\label{sec:prelims}
% \hast{We present the building blocks of our framework.
% We use EAU to describe the obligations of Markov Decision Processes (MDPs).
% We employ these descriptions in two tasks: model checking and policy improvement.
% The models that we check for conformance with an obligation are DAC-MDPs that are capable of representing an agent in a large, complex environment.
% To improve an MDPs conformance with an obligation, we cast the MDP as a parametric Markov Chain and find policy gradients to two ends. 
% First, to increase the probability with which the agent will perform the obligation.
% And second, to increase the performance of the agent on their original reward function.}

% \hahl{We describe DAC-MDPs, this paper's model for action under uncertainty.
% We then define Expected Act Utilitarianism, the logic for formalizing obligations of agents acting in (DAC-)MDPs.}

We introduce the Expected Act Utilitarian deontic logic for formalizing obligations of agents acting in stochastic environments.
We then draw a correspondence between the semantic frames of the logic, MDPs, and Bellman optimality.
And we discuss the policy gradients that are central to our policy search algorithms.

\subsection{Expected Act Utilitarianism}
\label{sec:eau}

\begin{figure}[t]
    \centering
    \includegraphics[width=0.75\linewidth]{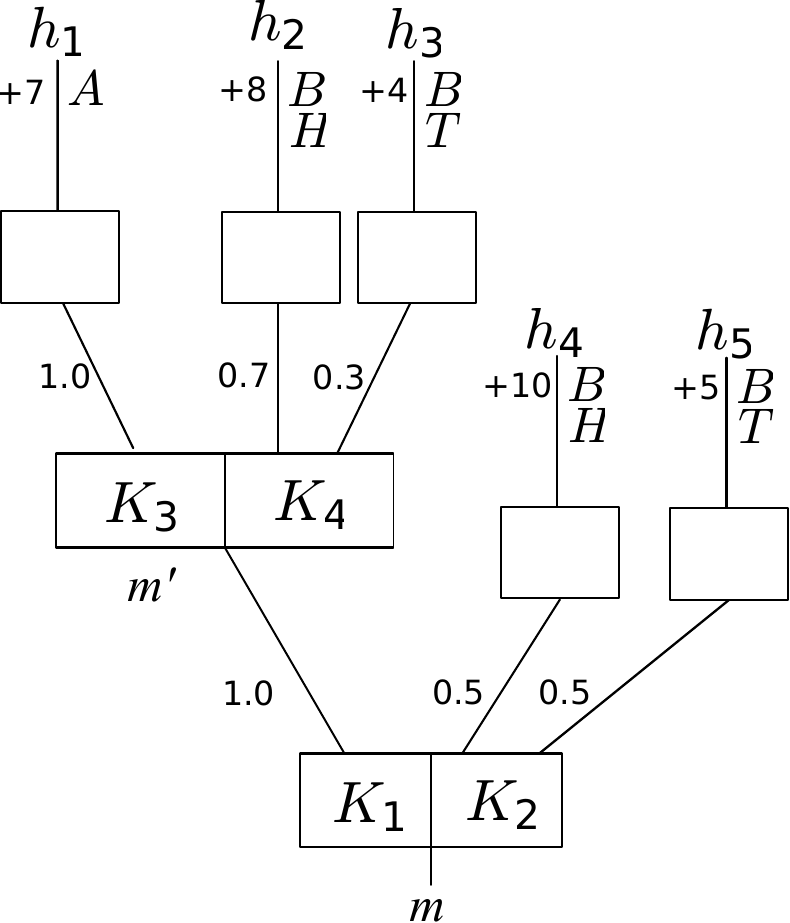}
    \caption{
        An EAU model for agent $\alpha$, showing moments $m < m'$ with histories $H_m = \{h_1, \dots, h_5\}$, and $H_{m'} = \{h_1, h_2, h_3\}$.
        The actions available in moment $m$ are $\Choiceam=\{K_1, K_2\}$, and in $m'$ are $\Choice{\alpha}{m'}=\{K_3, K_4\}$.
        Action $K_1 = \{h_1, h_2, h_3\}$, $K_2 = \{h_4, h_5\}$, $K_3 = \{h_1\}$, and $K_4 = \{h_2, h_3\}$.
        Each history is labeled with the formula(s) it satisfies, and its values $\Value(h)$; e.g., $h_1$ satisfies $A$ and has a value of 7.
        The probability of an action being able to effect a moment or history is also given; e.g. $\Proba(m' | m) = 1.0$, and $\Proba(h_2 | m') = 0.7$.
        The index $m/h_4 \models \estit{\alpha}{B}$ since $\Choiceam(h_4) = K_2 = \{h_4, h_5\}$, and both $h_4$ and $h_5$ satisfy $B$.
        However, $m/h_4 \not\models \estit{\alpha}{H}$ because $h_5$ does not satisfy $H$.
        Still, $m/h_4 \models \estit{\alpha}{P_{\geq 0.5}[H]}$ since $\Proba(h_4 | m) \geq 0.5$.
        And $m/h_2 \models \estit{\alpha}{P_{\leq 0.7}[H]}$ because $\Choiceam(h_2) = \{h_1, h_2, h_3\}$, $h_2$ is the only history among those that satisfies $H$ and $\Proba(h_2 | m) \leq 0.7$.
        The $Q(K_2) = 7.5$, while $Q(K_1) = 1.0 * \max \{Q(K_3), Q(K_4)\} = \max \{7.0, 6.8\} = 7.0$, so $\Optimalam = \{K_2\}$.
        Hence $m/h_4 \models \OEstit{\alpha}{B}$.
        Finally, $\Optimal{\alpha}{m'} = \{K_3\}$, so $m'/h_2 \not\models \OEstit{\alpha}{B}$.
    }
    \label{fig:model}
\end{figure}

Expected Act Utilitarianism, or EAU~\cite{shea2022eau}, uses PCTL~\cite{baier2008principles} to describe states of affairs in the world, and adds modalities to speak of action and obligation.
% as a tense logic, and reasons about optimal actions in terms of expected discounted reward.
Letting $\alpha$ be an agent from a finite set of agents $\Agent$, $\land$ and $\neg$ be Boolean conjunction and negation, and $\formula$ a PCTL formula, the syntax of EAU is defined by the following grammar,
\begin{equation*}
A \defeq \formula~|~\neg A~|~A\land A~|~~\estitaa~~|~~\OEstit{\alpha}{A}
\end{equation*}

\hahl{Intuitively, PCTL formula $\formula$ describes a state of affairs, such as $P_{\geq 0.9} \eventually g$: the probability of predicate $g$ eventually holding is at least 0.9. See \cite{baier2008principles} for details of PCTL.}
Formula $\estitaa$ says that $\alpha$ ``sees to it that'', or ensures, that $A$ is true, while $\OEstit{\alpha}{A}$ says that $\alpha$ \textit{ought to ensure} that $A$ is true.
For example, the formula $\OEstit{\alpha}{P_{\geq 0.75}[\neg \eventually playground]}$ specifies the obligation to avoid the playground in at least 75\% of possible executions.
The formula $\OEstit{\alpha}{P_{< 0.01}[\eventually checkpoint]}$ gives the obligation to eventually reach the checkpoint in less than 1\% of possible executions. 

Figure \ref{fig:model} illustrates all definitions in this section.
Formally, EAU formulas are interpreted over a \textit{branching time model} $\Model$. 
It is made of the following components: %stit tree \textit{Tree}, which models a non-deterministic execution of an MDP. Namely, for a stit tree,
\begin{itemize}
    \item A tree \textit{Tree}, whose vertices are called \textit{moments}, $m$, representing decision points of the agent. Two moments are related by a directed edge $mm'$ if $m'$  follows $m$. This is a partial order on moments, written as $m<m'$.
    \item the \textit{root} is moment `0', which precedes all other moments.
    \item $AP$ is a set of atomic propositions, and $v: V(Tree)\rightarrow 2^{AP}$ is a labeling function that assigns atomic propositions to each moment $m$.
    \item a \textit{history} $h$ is a linearly ordered, \yhl{possibly} infinite, set of moments --- a branch of the \textit{Tree}. We write $H_m$ for the set of histories that start at moment $m$
    \item $K$: an action available to an agent at a moment $m$. By identifying $K$ with the subset of histories in $H_m$ that are still possible after taking the action, we can consider that $K\subseteq H_m$. We write $\Choiceam$ for the set of actions confronting $\alpha$ at $m$, and $\Choiceam(h)$ to refer to those actions that contain history $h$.
    \item $\Proba(m' | m)$: the probability of agent $\alpha$ moving from $m$ to $m'$, assuming that the agent takes \yhl{some} action $K$ that leads to $m'$ (formally,  $K \subseteq H_m$ and $K \cap H_{m'} \neq \emptyset$).
    \item $\Value: h \rightarrow \Value(h)$: a function that assigns a real value - a utility - to a history.
\end{itemize}

The branching time model can represent the roll-out of an MDP, where moments are state visits, and probabilities are derived from transition probabilities.

An EAU formula holds (or not) at an index of evaluation in the model, which is a moment/history pair $m/h$.
The satisfaction relation is written $\Model, m/h \models A$, where it is always the case that $h \in H_m$.
The \textit{proposition} defined by the EAU statement $A$ at moment $m$ is the set of histories, starting at $m$, in which the statement holds:
\begin{equation}
    |A|_m^\Model \defeq \{h \in H_m \such \Model, m/h \models A\}
\end{equation}
When it is unambiguous, we drop $\Model$ from the notation.

For convenience, we write the probability with which an agent can execute a particular history $h$ from moment $m$ as $\Proba(h|m)$.
We can determine this value by taking the product of the probabilities $\Proba(m' | m)$ along history $h$ --- assuming that the agent always takes the action $K$ that history $h$ is a part of, and dropping $K$ from the notation.

The \textit{quality} of an action, $Q(K)$, is defined as:
\begin{equation}
    \label{eq:quality}
    Q(K) = \sum_{m' \in M'_{K}} \Proba(m' | m) \max_{K' \in \Choice{\alpha}{m'}} Q(K')
\end{equation}
where $M'_{K}$ is the set of moments that follow $m$ by taking action $K$.

An agent's set of optimal actions, then, can be defined as the action(s) with the best quality at the moment:
\begin{equation}
    \label{eq:optimalam}
\begin{aligned}
    \Optimalam \defeq \{K \in \Choiceam &\such \not\exists K' \in \Choiceam
    & \\ \textrm{s.t.}~Q(K)< Q(K')\}
\end{aligned}
\end{equation}

Now we can provide the formal semantics for two important concepts in EAU: \textit{agency} and \textit{obligation}.
In EAU, \textit{agency} is modeled by the `Chellas sees to it' operator $cstit$~\cite{chellas1968logical}. 
The agent $\alpha$ ``sees to it that'' the statement $A$ holds at index $m/h$ if and only if it takes an action $K$ such that $A$ holds in all histories  in $K$  - i.e., $K$ guarantees $A$ is true \cite{Horty01DLAgency}.
We write this as $\Model,m/h \models [\alpha\,cstit:A]$.

% \todo[inline]{note that the obligation at m is a consequence of the optial actions at m alone - no consideration is made to what is optial int he future. By cntrast, an optial policy (that maximizes utility) is typicaly followed forever -> disconnect which we addresss in section 4. \checkmark}
We model \textit{obligation} with the expected Ought.
The agent $\alpha$ ``ought to see to it that'' $A$ holds at index $m/h$ if and only if $A$ is guaranteed by every \textit{E-Optimal} action.
% is a necessary outcome of all of $\alpha$'s optimal actions at $m$.
In other words, an agent's obligations at $m$ are defined by the states of affairs $A$ guaranteed by taking its optimal actions at this moment.
To denote that an obligation is satisfied we write $\Model,m/h \models \OEstit{\alpha}{A}$.
\begin{definition}[Expected Ought]
	\label{def:ought semantics}
    With $\alpha$ an agent and $A$ an obligation in a model $\Model$, 
    \begin{equation}
        \label{eq:ought}
        \begin{aligned}
            \Model,m/h \models \OEstit{\alpha}{A} &\text{ iff } K \subseteq |A|_m^{\Model}\\
            &\quad \text{ for all } K \in \Optimalam
        \end{aligned}
    \end{equation}
\end{definition}

\yhl{Note that an obligation at moment $m$ is a function of the optimal actions at $m$ alone. No consideration is given to what is optimal in the future or what actions are taken in the future. By contrast, an optimal policy (that maximizes expected utility) is typically followed forever. To reason about an agent's obligations under the consideration that it will always act in accordance with its optimal policy, we introduce a new modality to the logic in Section \ref{sec:mc}.}

\textbf{From Markov processes to stit models.} 
In the context of Markov processes, a stit model can be thought of as a roll-out of the process.
A moment $m$ is assigned to each visit of each state in the roll-out, and the \textit{root} moment is the roll-out's first visit to the starting state.
Each action $a$ in the MDP has a corresponding action $K$ for each moment that action can be taken, and the probability $\Proba(m'|m)$ is the transition probability $T(s, a, s')$ for the states $s, s'$ associated with the moments $m, m'$, and the action $a$ associated with action $K$.
A history $h$ is a particular realization of an execution, and its $\Value$ would be its discounted sum of rewards.
Thus the $Q$\textit{-function} derived for the MDP is isomorphic to the $Q(K)$ value for an action in the \textit{Tree}, and so, as shown in \cite{shea2022eau}, an agent's optimal action $K \in \Optimalam$ \hahl{in the stit model} is its Bellman optimal action \hahl{in the MDP} \cite{Bellman1957markovian}.

% \todo[inline]{@Aayam please describe this}

\subsection{Property Gradients for Parametric Markov Chains}
\label{sec:pmc}
In the second part of this paper we will propose an algorithm for updating an optimal policy so that it satisfies the content of an obligation - that is, the PCTL formula $P_{\geq \rho} \phi$ that shows up in an obligation operator $\OEstit{\alpha}{P_{\geq \rho} \phi}$.
To do this we leverage recent work \cite{Badings2023prmc}, in which gradients with respect to transition probabilities are computed. 
Specifically, given a parametric Markov Chain (MC) whose transition probabilities are parameterized, and some function $f: S\rightarrow \mathbb{R}$ of the states which is obtained as the solution of a linear program, \cite{Badings2023prmc} show how the gradient of $f$ with respect to the parameters can be computed efficiently.
\yhl{In practice, the parameters on the transition probabilities represent a policy.}
\yhl{The function $f$ represents either the reward function (when we want to maximize expected utility), or the probability of satisfying the formula $\phi$ (when we want to maximize the likelihood of satisfaction).}

%=============================================
\section{Model-Checking Strategic Obligations}
\label{sec:mc}
We aim to extend EAU to describe what an agent should do given that it follows its optimal policy for all time. Following \cite{Horty01DLAgency}, we call these \textit{strategic obligations}.
This makes EAU more applicable to the RL paradigm, where the optimal policy is typically followed forever. 
We therefore extend EAU to speak of \textit{strategic obligations}, then present a model-checking algorithm to handle strategic obligations.

\subsection{Expected Strategic Oughts}
% \todo[inline]{first para a bit repetitive, can cut out if need space}
While EAU gives us the capability to specify and reason about an agent's obligations in stochastic environments, the obligations it defines are determined only by the agent's optimal action in the moment of evaluation.
When an agent follows a strategy however, it is natural to inquire about the implications that strategy has on its behavior (beyond its immediate impact).
This is especially pertinent when verifying RL systems, as they are expected to follow a learned policy.

To reason about an agent's obligations under a strategy, we introduce a strategic obligation in the manner of Horty's strategic ought \cite{Horty01DLAgency}, though a broader treatment of strategic modalities can be found in \cite{broersen2015using}.

% \todo[inline]{specific $\pi$ or generic symbol for strategy? re-work, including does pi map states or moments \checkmark}
A strategy (or policy, or schedule) $\pi$, is a mapping from \yhl{moments} to actions that determines what action an agent will take when it finds itself in a given state.
A strategic obligation, then, is an agent's obligation \yhl{to }the state of affairs brought about by an agent following its optimal strategy.
We begin with the strategic \textit{stit} modality: $\sstitaap$, which says that agent $\alpha$ \yhl{has some strategy $\pi$ that, if followed, ensures that $A$ is the case.}
To say that an agent has a strategic obligation we write $\OEsstit{\alpha}{A}{\pi}$.

We define a strategy $\pi$ as a mapping from moments $m$ in the stit tree to a subset $\pi(m)$ of the actions available at $m$.
The set of histories realizable by $\pi$ starting at $m$ is then 
\[H_{m,\pi} = \{h \in H_m | h \in \pi(m') ~\forall m' \in Tree \text{ s.t. } m'\geq m\}\]
Thus $h\in H_{m,\pi}$ iff $h$ is a possible evolution of the system if the agent follows $\pi$.

%We then say that an agent $\alpha$ ensures that $A$ holds by following strategy $\pi$ from index $m/h$ if and only if $H_{m,\pi} \subseteq |A|_m$. An agent strategically sees to it that $A$ is the case if there exists a strategy $\pi$ that ensures $A$ when it follows $\pi$.
\begin{definition}[Strategic stit]
    \label{def:sstit}
    In a stit model $\Model$,
    $\Model, m/h \models \sstitaap$
    iff there exists a policy $\pi$ such that
    $h \in H_{m,\pi} \text{ and } H_{m,\pi} \subseteq |A|_m^\Model$.
\end{definition}

To say that an agent has a strategic obligation we must return to the question of optimality.
We say that a strategy is optimal if all actions in it are optimal actions.
That is, $\pi$ is $\alpha$'s \textit{optimal strategy} if and only if $K \in \Optimalam$ for all $K \in \pi$, and for all $m$.
Then we can say that an agent $\alpha$ has an obligation to strategically ensure that $A$ holds if and only if all optimal strategies ensure $A$.
\begin{definition}[Expected Strategic Ought]
    \label{def:esought}
    In a stit model $\Model$,
    $\Model, m/h \models \OEsstit{\alpha}{A}{st}$
    iff $H_{m,\pi} \subseteq |A|_m^\Model$ for all optimal $\pi$
\end{definition}

% \todo[inline]{fix $\pi1$ to $\pi_1$, etc \checkmark}
To illustrate these operators, we turn to Figure \ref{fig:model}.
\yhl{Let us define $\pi_1$ as $\{m:K_1; m':K_3\}$, $\pi_2$ as $\{m:K_2; m':K_3\}$, and $\pi_3$ as $\{m:K_1, K_2; m':K_3, K_4\}$.}
We can say that $m/h_1 \models \sstit{\alpha}{A}{\pi}$ as $\pi_1(m) = \{K_1\}$, and $\pi_1(m') = \{K_3\}$, so $H_{m,\pi_1} = \{K_1\} \cap \{K_3\} = \{h_1\}$ and $|A|_m = \{h1\}$ --- \yhl{that is, there is a strategy ($\pi_1$) that guarantees $A$}.
The optimal action at $m$ is $K_2$, and the optimal action at $m'$ is $K_3$, so we can say that $\pi_2$ is the optimal strategy, and that $m/h_4 \models \OEsstit{\alpha}{B}{\pi2}$.
Finally, we note that since $H_{m,\pi_3} = \{h_1, h_2, h_3, h_4, h_5\}$, we can't say that $\pi_3$ guarantees any atomic proposition.

\yhl{If $\Value{(h_1)}$ were 10 instead of 7, then $\pi_1$ would be the optimal strategy. In this case, $m/h_1 \models \OEsstit{\alpha}{A}{\pi_1}$. In EAU \textit{without} strategic obligations, however, we cannot assume that the agent will continue to act optimally in the future. Thus $m/h_1 \not\models \OEstit{\alpha}{A}$.}

\subsection{Model Checker of Strategic Oughts}
To make the strategic EAU modalities a practical tool for the verification of RL agents, we developed and implemented a model checking algorithm for strategic obligations.
% \todo[inline]{following sentence not consistent with algo: algo takes an MDP, not a branching time model. So just say here the inputs are an MDP, bla and bla}
% The problem of strategic EAU model checking is: given an EAU branching time model $\Model$, determine whether $\Model, 0/h \models \OEsstit{\alpha}{\phi}{\pi}$ for some history $h \in H_0$, where $\phi$ is a formula in PCTL.
Our model checking algorithm takes as inputs an MDP $\Model$, and an obligation $\OEsstit{\alpha}{\phi}{\pi}$ where $\phi$ is a formula in PCTL.
The algorithm then determines if it is the case that $\Model \models \OEsstit{\alpha}{\phi}{\pi}$
By definition \ref{def:esought}, this model checking problem can be performed in two sequential steps: first, find the optimal strategy $\pi^*$, and second, determine if all histories consistent with $\pi^*$ satisfy $\phi$ .
% (i.e. $H_{0,\pi^*} \subseteq |\phi|_0^\Model$).
% \todo[inline]{this assumption was an issue for one of the reviewers. You must address that concern: either give a class of systems where the policy is unique, or expound, convincingly, on why this is not a terrible restriction. You say the agent will only follow one policy - so? Elaborate.}
\yhl{Our model checker assumes that the optimal policy $\pi^*$ is unique.}
To enforce this assumption we simply employ a tie-breaking rule, forcing the selection of only one optimal policy if multiple are available.
This is done for simplicity; if the existence of multiple optimal policies is critical then the algorithm could be modified to return true only if all optimal policies satisfy the obligation.
%If all strategy-consistent histories guarantee $A$, then we can say that $\Model$ has the obligation to strategically ensure $A$ at index $0/h$.
This algorithm is shown in Algorithm \ref{alg:model_check}, and experimental results of its performance are given in Section \ref{sec:experiments}.
% \todo[inline]{Something's off with the symbols in the algorithm, you use $B$, $A$ and $\phi$ to refer to the same thing (PCTL content)? This sort of stuff pisses off reviewers and if entirely avoidable.}
% \todo[inline]{in algo, instead of "PCTL component" just give the obligation statement, consistent with the text above.}

\begin{algorithm}
    \caption{Strategic EAU Model Checking}
    \label{alg:model_check}
    \begin{algorithmic}
    % \SetKwFunction{ValIt}{ValueIteration}
    % \SetKwFunction{Remodel}{RemoveAction}
    % \SetKwComment{Comment}{/* }{ */}
    \REQUIRE an MDP $G$, a state $s_0$ in $G$ to check from, an EAU obligation $\OEsstit{\alpha}{\phi}{\pi}$.
    \ENSURE a Boolean $\top$ if $\Model \models  \OEsstit{\alpha}{\phi}{st}$, otherwise $\bot$.
        \STATE $\pi^* \gets $ValueIteration$(G, s_0)$ \COMMENT{use value iteration to find optimal policy}
        % \todo[inline]{what does $K\in \pi$ mean? that $K$ never shows up in any state? or are you remving, in each state, those actions that are not assigned by $\pi$? needs clarification to avoid reviewer knocking us over this \checkmark}
        \FOR{$(s, a, s') \in T$}
            \IF{$a \not\in \pi^*(s)$}
                \STATE $G \gets $RemoveTransition$(G, (s, a, s'))$ \COMMENT{disable each state-action pair that isn't in the policy}
            \ENDIF
        \ENDFOR
        \STATE $valid \gets G \models \phi$ \COMMENT{call Storm to check PCTL formula $\phi$}
        \IF{$valid$}
            \RETURN $\top$ \COMMENT{the policy ensures $\phi$}
        \ELSE
            \RETURN $\bot$
        \ENDIF
    \end{algorithmic}
\end{algorithm}

%=============================================
\section{Policy Update to Satisfy Obligations}
\label{sec:policy update}

\begin{figure*}[ht]
    \centering
    % \vspace{-2cm}
    \begin{subfigure}[t]{0.32\textwidth}
        \includegraphics[width=\textwidth]{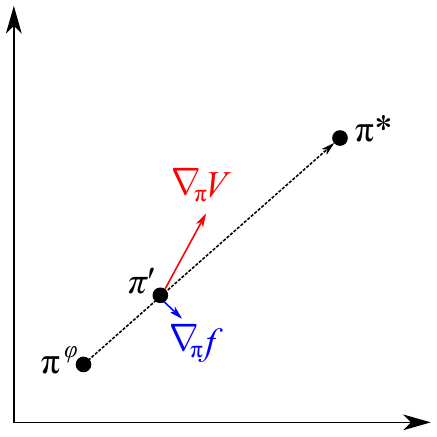}
        \caption{Line Search. Starting from $\pi^\varphi$, the policy moves along the line towards $\pi^*$. The policy on this line that maximizes utility while satisfying the probability threshold is returned.}
        \label{fig:projected-grad}
    \end{subfigure}
    \hspace{0.01\textwidth}
    \begin{subfigure}[t]{0.32\textwidth}
        \includegraphics[width=\textwidth]{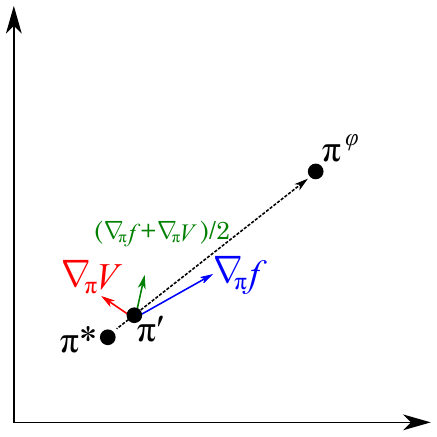}
        \caption{Average Gradient. The average (in green) of the utility gradient (red) and the probability gradient (blue) is used to update the policy.}
        \label{fig:avg-grad}
    \end{subfigure}
    \hspace{0.01\textwidth}
    \begin{subfigure}[t]{0.32\textwidth}
        \includegraphics[width=\textwidth]{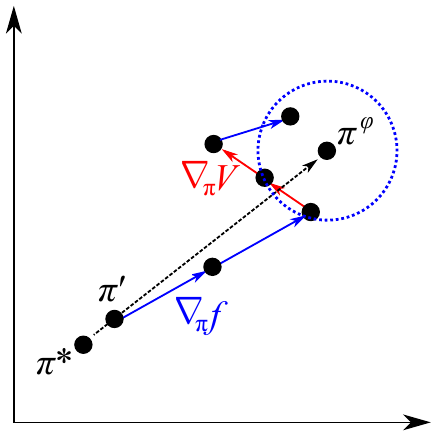}
        \caption{Alternating Gradient. If the obligation is not satisfied then the policy follows the probability gradient (blue). Otherwise, the policy follows the utility gradient (red).}
        \label{fig:alt-grad}
    \end{subfigure}
    \caption{Gradient-based constrained policy search methods. The gradient of the expected utility of a system with respect to its policy (the \textit{utility gradient}) is labeled $\nabla_\pi V$ and is red. This gradient points the policy in the direction that increases its expected utility. The gradient of the probability that the policy satisfies the specification with respect to its policy (the \textit{probability gradient}) is labeled $\nabla_\pi f$ and is blue. This gradient points the policy in the direction that increases the probability with which it satisfies the given obligation. The policy that maximizes this probability is labeled $\pi^\varphi$, and the policy that maximizes expected utility is labeled $\pi^*$.}
    \label{fig:grad-diagrams}
\end{figure*}

In this part of the paper, we move from model-checking an MDP against a strategic obligation, to policy search with strategic obligation constraints. 
Namely, the design team is given an obligation $\fullsstit$ as part of design requirements. 
The team comes up with a reward structure for the agent. 
Ideally, the reward function induces an optimal policy $\pi^*$ that also satisfies the obligation, but that is \textit{not guaranteed} since the reward might be balancing several requirements, and reward shaping is notoriously difficult. 
But the ethical obligations aren non-negotiable. 
We therefore ask: how can we modify the reward optimal $\pi^*$ to obtain a policy $\pi'$ such that the controlled system satisfies $P_{\geq \rho}\varphi$ while maintaining a high expected reward?

We do this by leveraging gradient computation for parametric MCs~\cite{Badings2023prmc}, which was described in Section~\ref{sec:prelims}. 
In our case, the parametric MC is the underlying MDP controlled by $\pi^*$.
The parameters of the MC are the probabilities of taking a given action in a given state, i.e. $\pi(a|s)$. 
The function $f$ to be optimized is the probability of the parametric MC satisfying $\varphi$: this probability is computed as the solution of a linear program~\cite{baier2008principles}. 
The gradient of $f$ relative to the policy parameters is computed by solving a system $E$ of linear equations~\cite[Sect. 4.1]{Badings2023prmc}.
We call this the \textit{probability gradient} $\nabla_\pi f$.
By doing gradient ascent on $f$ relative to the action probabilities, we can increase the probability of satisfaction, at the cost of veering away from the initial, reward-optimal policy $\pi^*$.
This process does not change the agent's reward function --- allowing us to continuously evaluate our modified policy on the reward function given by the design team.
Note that the probability bound can also be $\leq \rho$ with trivial changes to the algorithm.
Note also that we can move along the top $k$ largest gradient entries, but potentially decrease search effectiveness. 
The impacts of this are explored in Section \ref{sec:illustrative}.

\paragraph{Approaches to policy updates.}
\yhl{Given the probability gradients needed to improve the policy's probability to satisfy $\varphi$, there are many heuristics that can be taken to update the policy while maintaining a high reward.}

% \todo[inline]{you're assuming for MC that optimal policy is unique, so it doesn't make sense to try this... I suggest removing it \checkmark}
% $\bullet$~\textit{Constrained Gradients: }We may instead seek to maintain reward optimality by modifying the system of linear equations $E$, whose solution yields the gradients. 
% Namely, we may add to it the linear constraints $V(s)=V^*(s), s \in S$, where $V^*(s)$ is the pre-computed optimal value function, achievable by $\pi^*$. 
% Of course, this new constraint might render the problem infeasible: a move away from $\pi^*$ is likely to decrease the value of some state, and the search must recover from this violation.

$\bullet$~\textit{Line Search: }One approach is to do a line search over the line connecting $\pi^*$, the reward-optimal policy, and $\pi^\varphi$, the policy that satisfies $P_{\geq \rho}\varphi$ (and which can be obtained by, e.g., STORM \cite{hensel2022storm}). This line is depicted in Figure \ref{fig:projected-grad}. This update simplifies the search space, and is doable using a classical projected gradient.

$\bullet$~\textit{Average Gradient: }We can also use the policy gradient $\nabla_\pi V$ for expected utility to guide our policy updates (the \textit{utility gradient}). By following the average of this gradient and probability gradient $\nabla_\pi f$, we can allow a simple trade-off between these two objectives.
% See Figures \ref{fig:tradeoff} and \ref{fig:avg-grad}.

$\bullet$~\textit{Alternating Gradient: }Finally, we can use the probability threshold $\rho$ to check when we should be following the probability gradient $\nabla_\pi f$ vs. the utility gradient $\nabla_\pi V$. If the current policy's probability does not meet the threshold $\rho$, then it is updated by following the probability gradient. If the threshold \textit{is} met, then the utility gradient is followed. This allows the policy to improve its utility only after it has sufficiently improved its probability of satisfaction.

In the next section we explore these three heuristics.
% \todo[inline]{re-work figure 3 to have less empty space, and increase the font of the text so it's legible when printed.}

% \begin{figure}[ht]
%     \centering
%     \includegraphics[width=0.75\linewidth]{AnonymousSubmission/figs/projected_gradient.pdf}
%     \caption{The reward optimal policy $\pi^*$ (right), the satisfying policy $\pi^\varphi$ (left), and the updated policy $\pi'$ (center) in policy space. The gradient of the updated policy with respect to the expected utility of that policy (the \textit{utility gradient}) is labeled $\nabla_\pi V$ and is red. This gradient points the policy in the direction that increases its expected utility. The gradient of the updated policy with respect to the probability that the policy satisfies the specification $\varphi$ (the \textit{probability gradient}) is labeled $\nabla_\pi f$ and is blue. This gradient points the policy in the direction that increases the probability with which it satisfies the given specification.}
%     \label{fig:projected-grad}
% \end{figure}

%Depending on which approach we adopt, the final policy might not be reward-optimal. It is therefore technically incorrect to say that the system 

%=============================================

\section{Experiments}
\label{sec:experiments}
To demonstrate the performance of our algorithms we report on the results of experiments on constrained policy search and model checking.
\yhl{An implementation of our model checking algorithm and our policy update algorithm is included with code to run the following experiments at:} https://github.com/sabotagelab/formal-ethical-obligations.

\subsection{Illustrative Example}
\label{sec:illustrative}
% \todo[inline]{put them in the design process? meaning what? \checkmark}
%To \shl{put EAU specifications into the design process for an agent}\yhl{bring EAU from \textit{post-hoc} model-checking to policy synthesis}, we introduce a double gradient ascent method that modifies the agent's policy while remaining sensitive to the policy's performance.
We first apply our methods to the \yhl{``windy-drone''} system depicted in Figure \ref{fig:windydrone}.
\yhl{The ``windy-drone'' system represents a drone delivering a heart for transplant while battling high winds. The effects of the wind are represented by stochastic transitions in the system. The drone's objective (as represented by its reward function) is to reach the hospital as quickly as possible. However, in such high winds, we want to prevent the drone from flying bio-hazardous material over locations such as playgrounds.}

The algorithm is given the stochastic policy that maximizes the agent's expected reward, and aims to modify the policy to satisfy $\OEsstit{\alpha}{P_{ \geq \rho }[ \neg \LTLfinally\, \yhl{playground} ]}{\pi}$.
This obligation represents the requirement for the agent to \yhl{avoid the playground} with a probability greater than $\rho$.
The dynamics of this obligation are interesting as fulfilling it pushes the agent away from its optimal policy --- encouraging it to take a path that is almost twice as long in the best case.
In this system, $\rho$ can not be larger than 0.998, \yhl{and for the following experiments we set $\rho = 0.75$}.
In practice, this probability threshold should be based on a risk analysis, and should reflect the degree of risk that stakeholders are willing to accept.

\paragraph{Line Search experiment.}

\begin{figure}[t]
    \centering
    \includegraphics[width=\linewidth]{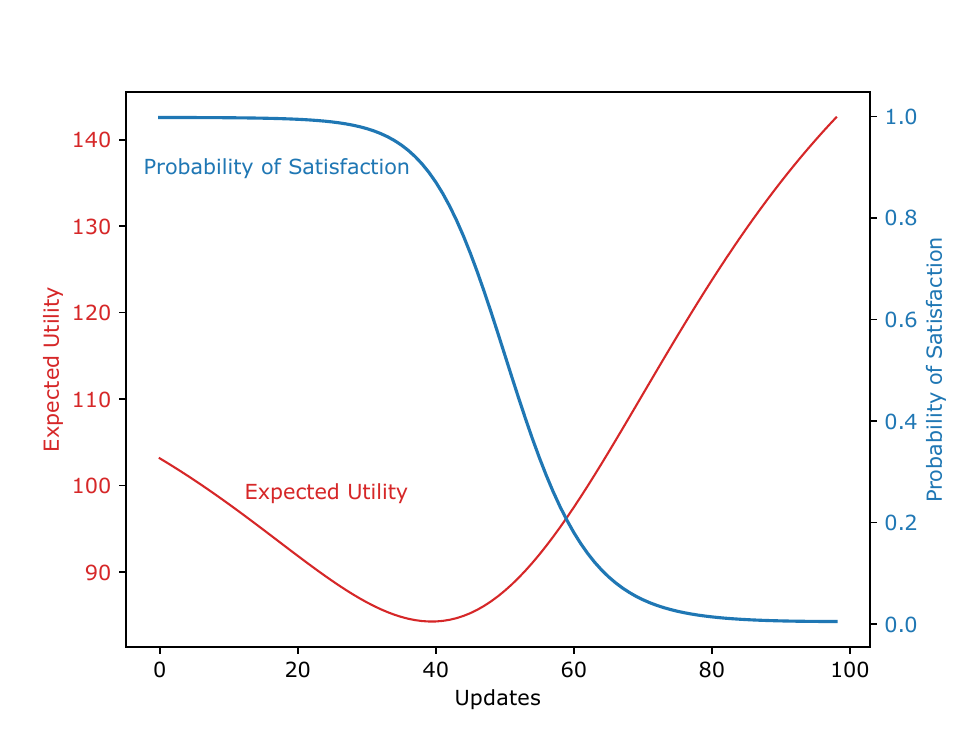}
    \caption{Line search experiment --- \yhl{Expected utility} (\yhl{red}) starts fairly high, but decreases as \yhl{satisfaction} probability (\yhl{blue}) decreases. It rebounds, however, as the policy moves closer to the reward optimal policy (at update 100). The \yhl{satisfaction} probability \yhl{begins near 1.0, but decreases to near 0.}}
    \label{fig:line-search}
\end{figure}

Our first experiment was on the performance of policies, on the line in policy space, between the the maximally satisfying policy $\pi^\varphi$ and the reward optimal policy $\pi^*$.
In this case we simply interpolate between the two policies --- taking 100 steps between $\pi^\varphi$ and $\pi^*$.
% \todo[inline]{your equation below (which is now commented out) says that you're multiplying each entry of $pi^*$ by $i$? But then suppose $\pi^*$ is 1 somewhere, then $1\cdot i$ is not even a probability... and $(1-i)\pi^\varphi$ is not even positive... the weights must be positive and sum to 1.  }
In Figure \ref{fig:line-search} we show the expected utility and probability of satisfaction of the policies on the line between $\pi^\varphi$ (update 0) and $\pi^*$ (update 100).
Each update $i$ on the x-axis describes the evaluation of a policy $\pi(i)$ defined by:
%\[\pi^\varphi (1-i) + \pi^* (i)\]
\[(1-i/100)\pi^\varphi + (i/100) \pi^*\]
As $i$ increases, the probability of satisfaction monotonically decreases.
The expected utility, however, initially decreases, but, around update 40, increases again as $\pi_i$ approaches $\pi^*$.
This non-monotonicity means that using a simple hill-climbing algorithm to find the maximum expected utility among policies on the line between two given policies will not necessarily return a policy with a global maximum expected utility.
We also note that the first update in this experiment to violate the probability threshold $\rho$ in our obligation of 0.75 is around $i=50$.
The policy that maximizes expected utility that is found before $\pi_{50}$ is $\pi_0$, which is $\pi^\varphi$.
% We can see that, in the interval where probability of satisfaction is above 0.75, the highest utility is at update 0.
% \todo[inline]{what non-convexity? the red line looks convex to me (but of course, the x-axis is nb of iterations so that's meaningless). what is the point(s)  you're trying to make? that there's no point on the line where both are maximized? That utility goes down and up?}
This figure lays the baseline for our other experiments in this section.

\begin{figure}[t]
    \centering
    \includegraphics[width=\linewidth]{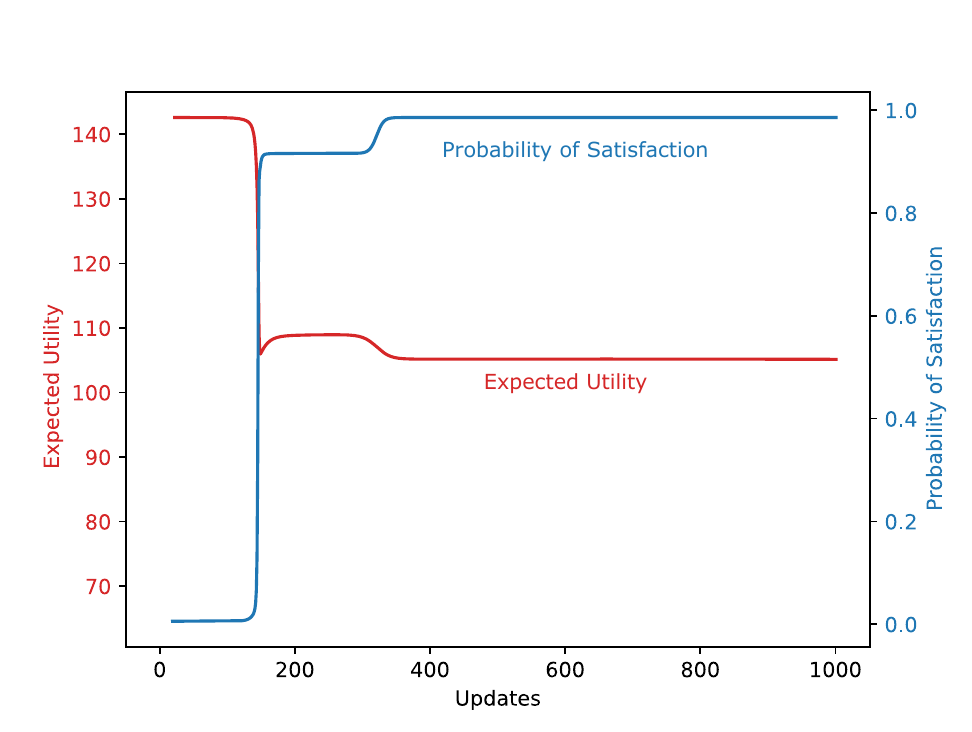}
    \caption{Average gradient experiment --- \yhl{Expected utility} (\yhl{red}) starts high, but decreases as \yhl{satisfaction} probability (\yhl{blue}) increases. The \yhl{satisfaction} probability \yhl{increases to be near 1.0, while the expected utility drops slightly below 110.}}
    \label{fig:tradeoff}
\end{figure}

\paragraph{Average Gradient experiment.}
% \todo[inline]{phosphorescent is a terrible color in the figure for this}
Our second experiment with this problem was on the trade-off between expected utility and probability of deontic satisfaction.
The algorithm does gradient ascent along $(\nabla_\pi V +\nabla_\pi f)/2$ - the average of the two gradients.
This method is depicted in Figure \ref{fig:avg-grad}, and the results are shown in Figure \ref{fig:tradeoff}.
\yhl{This allows the probability $f$ of satisfaction to rise to almost 1.0, while pulling the expected utility for the agent down below 110 from its maximum of 142.}
This suggests that there is room to increase utility again at the cost of some lowering of the probability of satisfaction (while maintaining $f \geq \rho$). 
Further, this approach gives no guarantee that $f$ will be raised above $\rho$. 
To address these two issues, we perform an experiment using the alternating gradient approach.

% \begin{figure}[ht]
%     \centering
%     \includegraphics[width=.75\linewidth]{AnonymousSubmission/figs/projected_gradient_avg.pdf}
%     \caption{Average gradient updates --- The average (in green) of the utility gradient (red) and the probability gradient (blue) is used to update the policy.}
%     \label{fig:avg-grad}
% \end{figure}

\begin{figure}[ht]
    \centering
    \includegraphics[width=\linewidth]{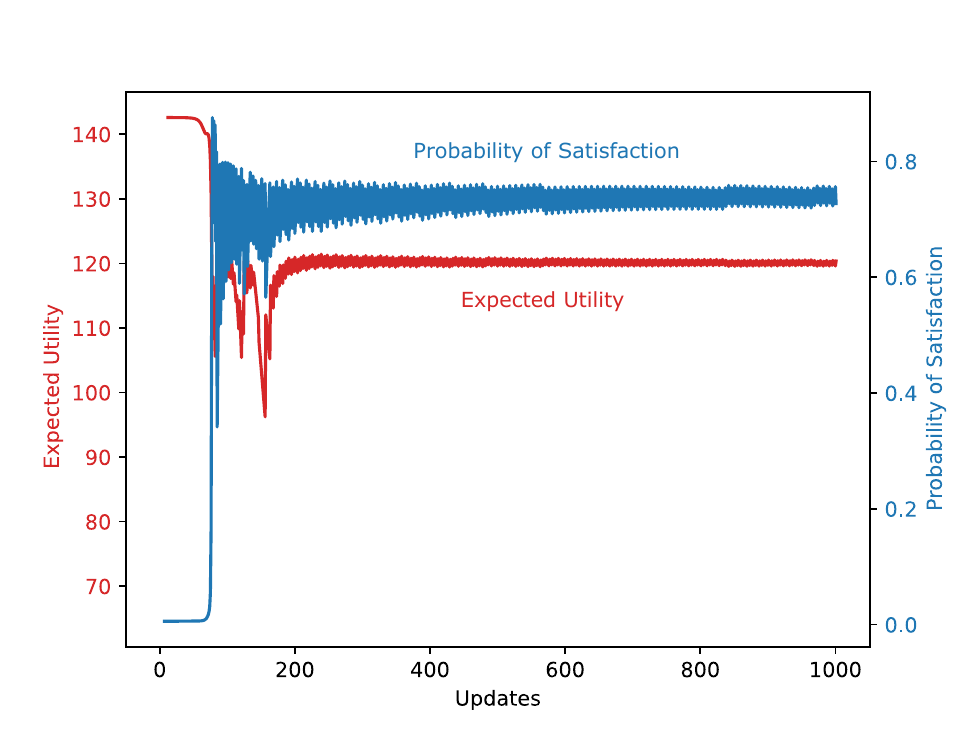}
    \caption{Alternating gradient experiment --- \yhl{Expected utility} (\yhl{red}) starts high, and \yhl{decreases some as the satisfaction probability (blue) increases}. Unlike in Figure \ref{fig:tradeoff}, the focus on the increase in probability of \yhl{satisfaction} until it reaches the threshold allows it to reach 0.75, and maintain a better performance around 120 --- an increase of almost 10\% over that of the average gradient experiment.}
    \label{fig:threshold}
\end{figure}

\paragraph{Alternating gradient experiment.}
Our third experiment uses the alternating gradient method depicted in Figure \ref{fig:alt-grad} where the policy is updated using the gradient that improves its satisfaction probability when below a probability threshold, and otherwise uses the utility gradient.
The performance of this method is shown in Figure \ref{fig:threshold} for a threshold $\rho=0.75$.
\yhl{Here, the probability of \yhl{satisfaction} comes to oscillate around the threshold, allowing the \yhl{expected utility} of the policy to rise to almost 120 --- an increase of almost 10\% over the line search and average gradient methods while maintaining satisfaction.}
\yhl{This method ensures that the satisfaction probability $f$ is near the set threshold, or is otherwise as high as possible. Only when $f$ is above the threshold does this method improve the agent's expected utility. The alternating gradient method is also less computationally demanding than the average gradient method, as each iteration only requires the calculation of one gradient.}

\begin{figure}[ht]
    \centering
    \includegraphics[width=\linewidth]{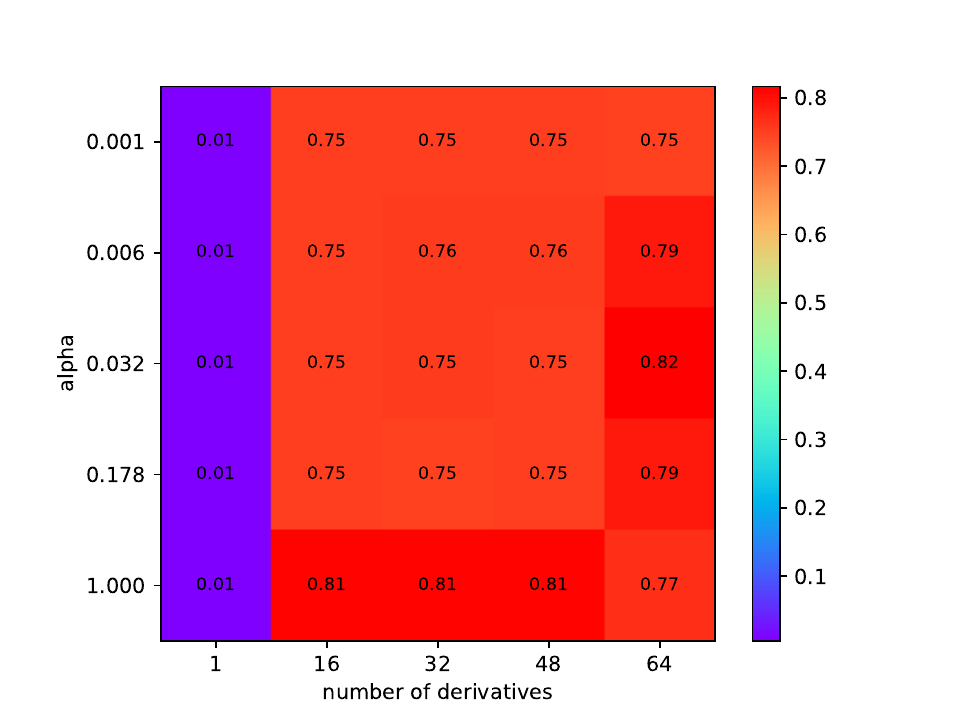}
    \caption{Alternating gradient experiment --- \yhl{Satisfaction} probability as a function of learning rate $\alpha$ and number of derivatives used in the gradient update. (Colors in digital version).}
    \label{fig:prob_mat}
\end{figure}

\begin{figure}[ht]
    \centering
    \includegraphics[width=\linewidth]{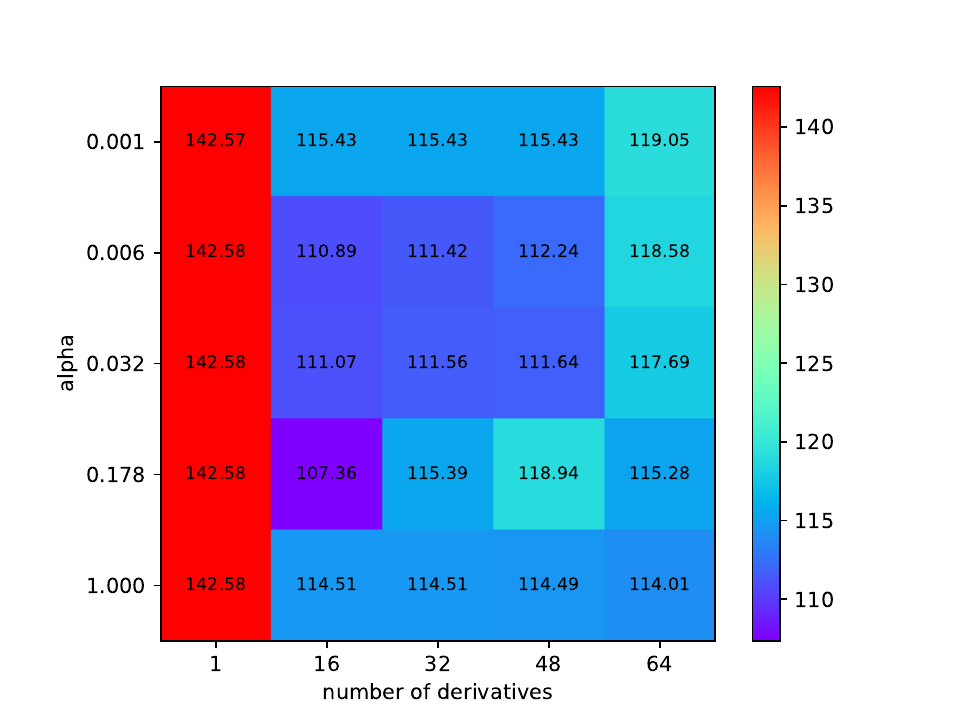}
    \caption{Alternating gradient experiment --- \yhl{Expected utility} as a function of learning rate $\alpha$ and number of derivatives used in the gradient update. (Colors in digital version).}
    \label{fig:val_mat}
\end{figure}

\paragraph{Hyperparameter choice.}
These experiments used a learning rate of 1.0, and all 64 of the system's derivatives \yhl{(4 for each of the 16 non-absorbing states)}.
However, choosing a learning rate, or number of derivatives is not straightforward.
To demonstrate, we performed a small grid search over these hyperparameters.
As seen in Figures \ref{fig:prob_mat} and \ref{fig:val_mat}, \yhl{the search for satisfying policies is robust to most choices of learning rate and number of partial derivatives.}
\yhl{It appears that only a few of the system's derivatives are required to get a good result. This is encouraging, as larger MDPs will require more time to solve, but may be solved without having the calculate a gradient value for every decision variable.}
\yhl{Using all of the derivatives, however, yields the highest final probability of satisfying the given specification. And, for small learning rates, it seems to lead to the best expected utility of those policies that satisfy the specification. At higher learning rates, using 48 of the 64 available gradients leads to a policy that satisfies the requirement, and performs slightly better than a policy found using all of the gradients.}

\subsection{Policy Optimization With Exploration}

We also sought to test if our method could perform well when rewards were not known \textit{a priori}.
To do this, we guessed the rewards at each state as an arbitrary value, and updated those values as the agent explored its environment.
This gave us access to an approximation of the utility gradient $\widetilde\nabla_\pi V$ which we can use to increase the expected reward of our policy.
% \todo{use \\widetilde everywhere}

However, if we want to ensure safety while exploring, then we need a way to prevent the agent from taking unsafe actions.
To this end we implement a PCTL \textit{shield} \cite{alshiekh2018safe} that prevents the agent from taking any action that would violate the content of a given obligation.

With the shield in place we can allow the agent to explore while we update its policy towards satisfaction of a given obligation.
Then, once we have found a safe policy, we can follow the approximate utility gradient that is based on our observations so far.
This is how we implement our alternating gradient method to allow for exploration.

To test if this method is effective across a broader range of environments we randomly generated 24 12-by-12 gridworlds.
Each had 10 impassible cells, 10 ``pits'' that would assign a reward of -10 and ends a run, 10 ``coins'' that would assign a reward of +5, and a goal state that would assign a reward of +10 and ends a run.
These cells would be randomly placed on the grid, and the agent would always start at the bottom-left of the grid.
The agent is initiated with a random policy, and may explore the environment $\epsilon$-greedily for 100 steps.
The agent's memory of reward values updates after every run, and the updated knowledge is used when calculating the current utility gradient.
The agent's policy is updated after every run as well using a learning rate of 0.01.

The obligation given to the agent is \[\OEsstit{\alpha}{P_{ > 0.75 }[ \LTLglobally\, \neg coin ]}{\pi}\]
That is, the agent should ensure that, with probability greater than 0.75, it should never enter a ``coin'' state.
This obligation was chosen because it interferes with the agent's ability to maximize its utility --- forcing $\pi^*$ and $\pi^\varphi$ to be different policies.

\begin{figure}[ht]
    \centering
    \includegraphics[width=0.9\linewidth]{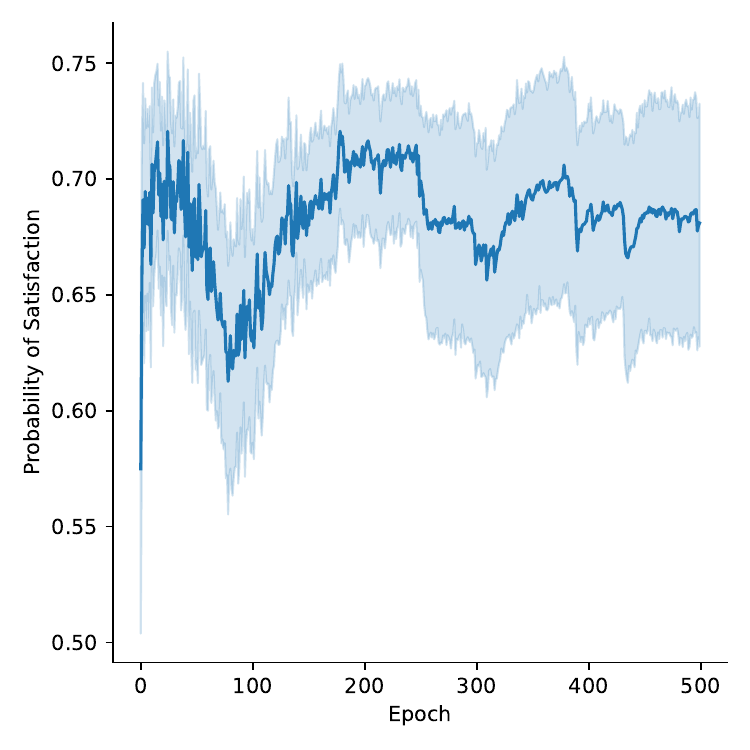}
    \caption{Alternating Gradient with Shield: Average probability of satisfaction as the agent explores and discovers reward values. The 80\% confidence interval is indicated by the shaded area. The probability increases as the probability gradient updates are applied, and then levels out below the threshold of 0.70.}
    \label{fig:apx-grad-sat}
\end{figure}

\begin{figure}[ht]
    \centering
    \includegraphics[width=0.9\linewidth]{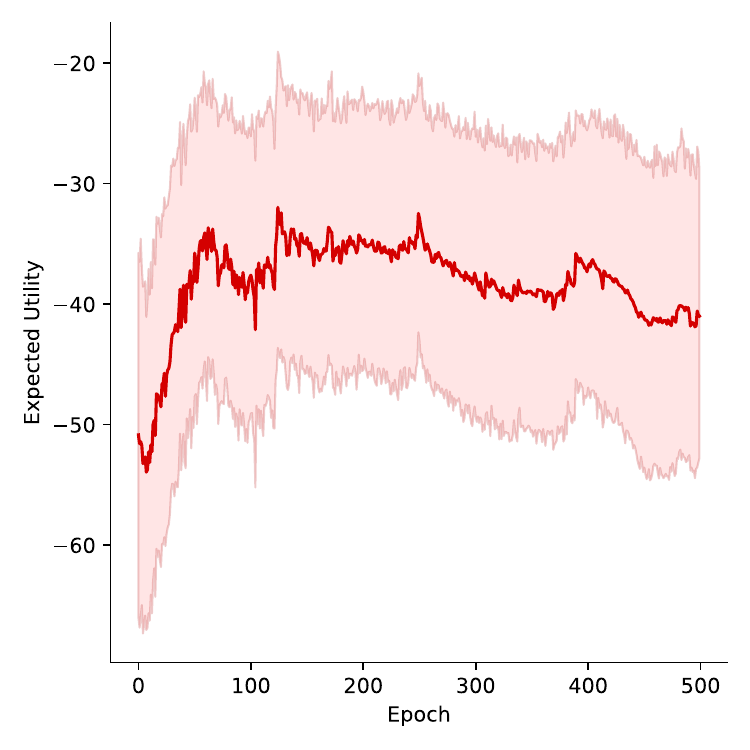}
    \caption{Alternating Gradient with Shield: Average expected utility as the agent explores and discovers reward values. The 80\% confidence interval is indicated by the shaded area. The utility increases from the random policy start, but decreases as the policy updates prevent it from gaining rewards from ``coins''. The width of the confidence interval is largely due to the variance in what utility our random environments allow.}
    \label{fig:apx-grad-val}
\end{figure}

The performance of this method is shown in Figures \ref{fig:apx-grad-sat} and \ref{fig:apx-grad-val}.
These show that as time goes on, the agent approaches the threshold of probability satisfaction (0.75), at which point the expected utility levels out.
We would expect the utility to decrease as satisfaction probability increases since the obligation prohibits the agent from collecting rewards from the ``coin'' states very often.

\begin{figure}[t]
    \centering
    \includegraphics[width=0.9\linewidth]{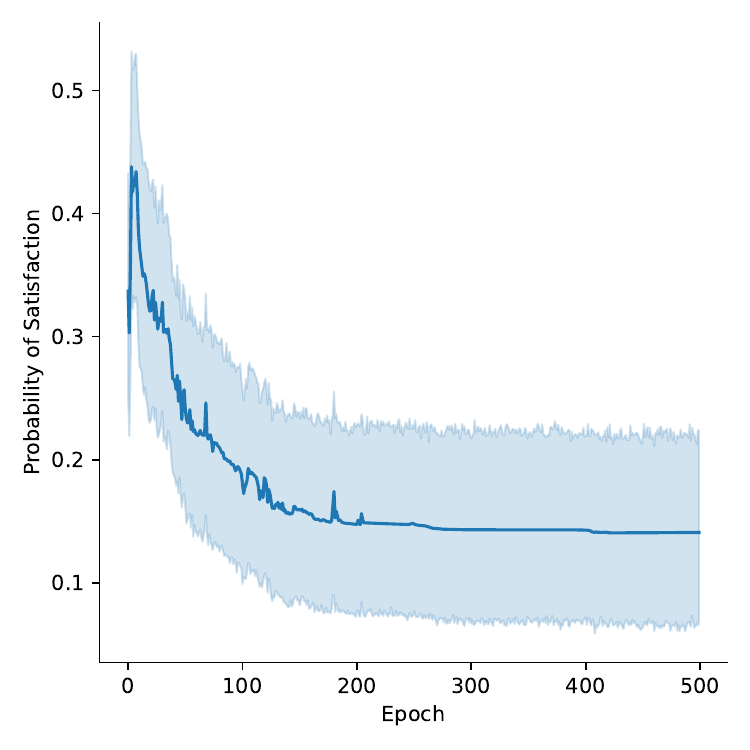}
    \caption{Utility Gradient with Shield: --- Average probability of satisfaction as the agent explores and discovers reward values. The 80\% confidence interval is indicated by the shaded area. Because the policy updates do not include the probability gradient $\nabla_\pi f$, the agent never learns to avoid the states that it is shielded from visiting, and so the probability of satisfaction does not increase.}
    \label{fig:shield-only}
\end{figure}

We also show that, for certain initializations of the rewards, shielding alone is insufficient to prevent the agent from learning an unsafe policy --- thereby requiring the shield to be used at execution time as well.
Figure \ref{fig:shield-only} shows that the satisfaction probability decreases, even though the agent never takes an action that would violate its obligations.
When shielded and following only the approximate utility gradient for its policy updates, the agent does manage to find an average reward near zero, but its probability of satisfying the obligation if the shield is removed is reduced below 0.2 on average.
By learning a safe policy using the alternating gradients method instead, we could remove the shield at runtime - saving on computational requirements.

\subsection{Large Model-Checking Experiment: Cartpole}
\label{sec:cartpole}
This experiment illustrates the execution of our model-checker on a large MDP  - specifically, on an MDP modeling the cartpole system.
The cartpole system consists of a pole attached by an un-actuated joint to a cart, which moves along a frictionless track. The system is controlled by applying a force of either positive or negative magnitude to the cart, with the objective of keeping the pole balanced upright without the cart running off the track. A tabular approximation of such a system can be defined by a DAC-MDP compiled on a set of trajectories collected by a random policy in the original MDP \cite{shrestha2020deepaveragers}. The size of the DAC-MDP can be adjusted based on the data size, fan-out, and fan-in size of the compiled MDP. Fan-out is controlled by the number of candidate actions for each state, and fan-in by the number of neighbors used to compile the DAC-MDP.

The MDP we retrieved had 50,000 states, 15 actions at each state, and 5 transitions per action, for over 3,000,000 transitions.
We formulated 20 PCTL formulas $\varphi_1,\ldots,\varphi_{20}$ to check as both strategic stit statements $\sstit{\alpha}{\varphi_k}{\pi}$, and as strategic obligations $\OEsstit{\alpha}{\varphi_k}{\pi}$.
We labeled each state with the quintile of the angle of the pole in that state $aq0 \dots aq4$, and with the quintile of the $x$-position of the cart $xq0 \dots xq4$.
The time it took to check a subset of these formulas is given in Table \ref{tab:check-times}.
We found that the time to model check strategic stit statements were consistently completed between 47 and 60 seconds.
All times were measured on a system with 16 GiB of RAM and an Intel i7-2620M CPU at 2.7 GHz.
Strategic obligations, with times near 25 seconds, were checked much more quickly; likely thanks to the induced Markov chain's smaller size - weighing in at fewer than 200,000 transitions.

\begin{table}[t!]
    \centering
    \caption{7 of the 20 formulas tested on the MDP with the amount of time, in seconds, that the model checking task took to complete. \yhl{The \textit{s-stit} time measures the time to check $\Model \models \sstit{\alpha}{\varphi_k}{\pi}$. The \textit{ought} time records the time to check $\Model \models \OEsstit{\alpha}{\varphi_k}{\pi}$.}}
    \label{tab:check-times}
    \begin{tabular}{l|l|l|l}
        & \textit{formula}                                & \textit{s-stit}  (s) & \textit{ought}  (s) \\ \hline
        $\varphi_1$ & $P_{ >= 0.2 }[ \LTLfinally\, ( aq0 | aq4 ) ]$     & 47.16            & 21.03          \\
        $\varphi_2$ & $P_{ >= 0.00001 }[ \LTLfinally\, ( aq0 | aq4 ) ]$ & 47.10            & 20.73          \\
        $\varphi_3$ & $P_{ >= 0.1 }[ \LTLglobally\, aq2 ]$                 & 55.42               & 24.63                \\
        $\varphi_4$ & $P_{ < 0.7 }[ \LTLglobally\, aq2 ]$                  & 47.83             & 20.80                \\
        $\varphi_5$ & $P_{ < 0.7 }[ \LTLfinally\, xq0 ]$                  & 56.60               & 20.92          \\
        $\varphi_6$ & $P_{ >= 0.7 }[ \LTLfinally\, xq0 ]$                 & 63.38               & 24.97                \\
        $\varphi_7$ & $P_{ > 0.7 }[ \LTLfinally\, xq0 ]$                  & 56.29               & 24.94                \\              
    \end{tabular}
\end{table}

\subsection{Obligation Implication}
% \todo[inline]{this section uses F for eventaually, use diamond like before \checkmark}
To demonstrate our policy optimization procedure on a more complicated form of obligation we show how to find a policy that satisfies an obligation with an implication within it.

\yhl{In the ``windy-drone'' system, we might consider allowing the drone to fly over a playground if it visits the checkpoint --- where its cargo is checked for safety and security.}
\yhl{We can model the obligation of the drone to fly over the playground (and thus take a shorter path to the hospital) if it visits the checkpoint as $\OEsstit{\alpha}{P_{\geq \beta}[\LTLfinally\, playground] \implies P_{\geq \gamma}[\LTLfinally\, checkpoint]}{\pi}$.}
\yhl{This is equivalent to saying that the drone should avoid the playground, or it should visit the checkpoint.}
\yhl{Thus, we can synthesize a policy that avoids the playground, and another policy that visits the checkpoint, and take the better performing of the two as the optimal satisfying policy.}

\yhl{In our experiments, we set $\beta$ to 0.75, and $\gamma$ to 0.9.}
\yhl{As shown in Figure \ref{fig:threshold}, a policy that avoids the playground results in an expected utility just below 120, while we found that visiting the checkpoint results in an expected utility of less than 40. Thus we take the former policy and complete the search.}\footnote{\yhl{The poor performance of the policy that visits the checkpoint is due to the Markov property of the MDP. If the policy could change \textit{after} it visited the checkpoint to allow the drone to take the faster northern route, then visiting the checkpoint would be the better-performing policy.}}

\subsection{Contrary-to-Duty Obligation}
As mentioned, a key strength of a deontic logic is the ability to distinguish between what ought to be the case (the obligation) and what actually is the case, and to reason over the divergence between these two.

A \textit{contrary-to-duty} (CTD) obligation is an obligation that enters into force in case a primary obligation (the duty) is violated: e.g. if the agent ought to ensure that the medicine cabinet is full (the primary obligation), but it isn't (the violation), then the agent ought to ensure that next an order is placed (the CTD). 
This has the following general structure: If $\alpha$ ought to ensure $A$ but $A$ does not happen, then $\alpha$ ought to ensure $B$ next. (Other structures are possible, notably using conditional obligations)
\begin{equation}
    \OEsstit{\alpha}{A}{\pi} \land \neg A \implies \OEsstit{\alpha}{\LTLnext B}{\pi}  
\end{equation}
We model-check such a formula on the \yhl{``windy-drone''} system depicted in Figure \ref{fig:windydrone}.
% \shl{By setting a high reward on state 10 (which can not be accessed after leaving), the agent will have the obligation $\OEsstit{\alpha}{\LTLnext P_{\geq 0.75} [ name=10 ]}{\pi}$.}
% \shl{However, in the case that the agent slips out of the start state, it will inherit the new obligation $\OEsstit{\alpha}{\LTLeventually P_{\geq 0.6} [ name=14 ]}{\pi}$.}
\yhl{By setting a high reward on the checkpoint state, the agent will have the duty $\OEsstit{\alpha}{\LTLnext P_{\geq 0.7} [checkpoint]}{\pi}$.}
\yhl{However, in the case that the agent slips north towards the playground (the violation), it will inherit the new CTD obligation to return to the start state so that it might make a second attempt at the checkpoint: $\OEsstit{\alpha}{\LTLnext P_{\geq 0.6} [ start ]}{\pi}$.}
\yhl{In the CTD structure we have:}
\begin{equation}
\begin{aligned}
    \OEsstit{\alpha}{\LTLnext P_{\geq 0.7} [checkpoint]}{\pi} \land \LTLnext north \\
    \implies \OEsstit{\alpha}{\LTLnext \LTLnext P_{\geq 0.6} [ start ]}{\pi}  
\end{aligned}
\end{equation}
\yhl{Verifying that our agent has this obligation allows us to determine how it is expected to behave when it enters a less-than-ideal state.}

We checked this contrary-to-duty obligations by forcing the agent to move north, and then verifying the truth of the CTD obligation from that state.
More generally, we check CTD obligations by looking at the successor states that would indicate a violation of the primary obligation and then verify the truth of the CTD obligations from those states.
% \todo[inline]{"This is checked" or we checked it? Use active voice, clarify what you did \checkmark}
%=================================================================

\section{Conclusions}
\label{sec:conclusion}
Our modalities for strategic agency and strategic obligation give us the expressive power to reason about a large class of reinforcement learning agents.
We presented an algorithm for model-checking that an MDP, equipped with a policy, has the right obligations captured in deontic logic.
We introduced a new kind of constrained MDP problem where reward maximization is constrained by an obligation.
We also provided a way to modify a policy so that it meets such an obligation --- without needing to toy with reward functions.

We hope that these algorithms will aid system designers in specifying normative constraints, checking their systems against those constraints, and refining their systems to meet their constraints.

\yhl{In future work we're interested in extending our policy update algorithm beyond safety properties to support a larger class of PCTL. We're also interested in managing the trade-offs between following the utility gradient and the probability gradient, perhaps by using a linear weighting function, or different learning rates for each. Further, we'd like to try constraining the policy search with a maximum divergence from the initial policy.}

\clearpage

\bibliography{aaai24}

\end{document}